\documentclass[lettersize,journal]{IEEEtran}

\usepackage{amsmath, amsfonts}
\usepackage{algorithm}
\usepackage{algorithmicx}
\usepackage{algpseudocode}
\usepackage{array}
\usepackage{subcaption}
\usepackage{booktabs}
\usepackage{textcomp}
\usepackage{stfloats}
\usepackage{url}
\usepackage{verbatim}
\usepackage{graphicx}
\usepackage{cite}
\title{Beyond Winning: Margin of Victory Relative to Expectation Unlocks Accurate Skill Ratings}

\author{Shivam~Shorewala, 
        Zihao~Yang%
\thanks{The authors are with Rimble Inc. (e-mail: \{shivam, zihao\}@rimble.io).}}

\begin{document}

\maketitle

\begin{abstract}
Knowledge of accurate relative skills in any competitive system is essential, but foundational approaches such as ELO discard extremely relevant performance data by concentrating exclusively on binary outcomes. While margin of victory (MOV) extensions exist, they often lack a definitive method for incorporating this information. We introduce Margin of Victory Differential Analysis (MOVDA), a framework that enhances traditional rating systems by using the deviation between the true MOV and a \textit{modeled expectation}. MOVDA learns a domain-specific, non-linear function (a scaled hyperbolic tangent that captures saturation effects and home advantage) to predict expected MOV based on rating differentials. Crucially, the \textit{difference} between the true and expected MOV provides a subtle and weighted signal for rating updates, highlighting informative deviations in all levels of contests. Extensive experiments on professional NBA basketball data (from 2013 to 2023, with 13,619 games) show that MOVDA significantly outperforms standard ELO and Bayesian baselines. MOVDA reduces Brier score prediction error by 1.54\% compared to TrueSkill, increases outcome accuracy by 0.58\%, and most importantly accelerates rating convergence by 13.5\%, while maintaining the computational efficiency of the original ELO updates. MOVDA offers a theoretically motivated, empirically superior, and computationally lean approach to integrating performance magnitude into skill rating for competitive environments like the NBA.
\end{abstract}


\section{Introduction}
\label{sec:introduction}

Quantifying skill and predicting outcomes in competitive scenarios, such as professional sports leagues \cite{stefani2011measurement} and multiplayer online games \cite{herbrich2006trueskill} to evaluating AI agents \cite{coulom2008whole} is a fundamental challenge with broad implications. Rating systems aim to solve this by inferring latent skill from observed performances. The ELO system \cite{elo1978rating}is one of the most praised tools for its elegant simplicity and interpretability. However, the ELO system's reliance on only binary win-loss results acts as a choke point. Standard ELO treats close call wins the same as blowout wins and thus ignores information about the magnitude of performance \cite{langville2012plays}. Losing this information creates large gaps in predictive accuracy and causes ratings to converge slower to true skill levels, especially as ratings evolve in dynamic environments \cite{hvattum2010using}.

Recognizing this deficiency, previous works have explored incorporating the margin of victory (MOV). These existing attempts often rely on ad-hoc heuristics, such as directly scaling ELO's K-factor by the raw MOV \cite{glickman1995adaptive, gill2021development}. These approaches lack a systematic way to determine how much margin is expected given the competitors' relative skill, rendering the raw MOV a potentially noisy and uncalibrated signal. Moreover, it still leads to an increase in the winning team's ELO even if the victory was not decisive, as their initial ELO difference might have indicated. At the same time, sophisticated Bayesian systems like Glicko-2 \cite{glickman1999parameter} or TrueSkill \cite{herbrich2006trueskill} model uncertainty effectively but primarily focus on outcome probabilities and often have significant computational overhead, typically do not explicitly model the expected performance margin itself. Therefore, a key gap persists: a method that integrates MOV information within an efficient ELO-like framework, based on a principled, learned model of expected performance margin.

This paper introduces Margin of Victory Differential Analysis (MOVDA), a unique approach designed to effectively fill this gap. MOVDA is grounded in the principle of comparing observed performance against a learned and context-aware expectation. We first model the expected MOV ($E_{MOV}$) as a non-linear function (specifically, a scaled hyperbolic tangent) of the pre-match rating difference ($\Delta R$), incorporating domain-specific factors such as home advantage. This function is calibrated on historical data. The core insight is to use the resulting differential, $\Delta_{MOV} = T_{MOV} - E_{MOV}$ (where $T_{MOV}$ is the true observed margin), as an extremely rich, informative signal for refining rating updates.

Our contributions are:
\begin{enumerate}
  \item \textbf{Principled Expected MOV Modeling:} We suggest and validate a theoretically motivated functional form (scaled hyperbolic tangent) to model expected MOV. This captures non-linearity based on rating differentials and different contexts.
  \item \textbf{MOV Differential Rating Update:} We develop an innovative ELO update rule that adaptively utilizes the MOV differential ($\Delta_{MOV}$), to provide a signal beyond the binary outcomes while importantly retaining the core ELO structure.
  \item \textbf{Significant Empirical Improvement on NBA Data:} We then demonstrate through extensive experiments on 13,619 professional NBA games (from 2013 to 2023) that MOVDA significantly outperforms standard ELO and other baselines like TrueSkill. Specifically, MOVDA achieves a 1.54\% lower Brier score (indicating better probabilistic predictions), 0.58\% higher outcome accuracy, and most importantly a 13.5\% convergence of ratings, allowing for extremely fast adaptation to changes in teams' abilities while keeping the same computational efficiency as the original ELO.
\end{enumerate}

\noindent
The structure of this paper is as follows. Section~\ref{sec:related_work} situates our work in the relevant literature on rating systems within a wider context. Section~\ref{sec:preliminaries} formalizes the existing ELO framework to provide a benchmark. Section~\ref{sec:methodology} introduces our contribution, our MOVDA framework, and outlines both its theoretical foundations and practical implications. Section~\ref{sec:experiments} offers a complete empirical evaluation. Section~\ref{sec:discussion} considers broader implications and acknowledges key limitations. Finally, Section~\ref{sec:conclusion} concludes the paper and proposes future potential directions.

\section{Related Work}

\label{sec:related_work}
Our work builds upon a substantial body of research on the design of rating systems. We group relevant literature into three primary pillars: (i) foundational ELO-based models, (ii) methods that incorporate margin of victory information, and (iii) probabilistic and Bayesian extensions.

\subsection{Foundational Rating Systems}
The ELO framework~\cite{elo1978rating} has long been a dominant method for competitive ranking systems, across domains from chess to esports. The ELO model, fundamentally, is a model of win probability as a logistic function of the rating differential and it updates a rating linearly based on observed outcomes. The ELO system has been successfully applied across many domains because its practical applicability, ease of interpretation, and computational efficiency.

Subsequent work has proposed numerous extensions the base ELO model, such as adaptive scaling constants ($c$), dynamic K-factors~\cite{glickman1995adaptive}, and context-aware adjustments. Theoretical treatments often recast ELO as a maximum likelihood estimator under parametric assumptions~\cite{glickman1999parameter}, or relate it to the Bradley–Terry model for paired comparisons. The standard ELO is binary, simply limited to win/loss results, and and does not account for the margin or quality of performance in its updates.

\subsection{Incorporating Margin of Victory}

It is now substantial evidence that margin of victory (MOV) has a significant informative signal. Early approaches incorporate the MOV via rule-based adjustments to the ELO update. A common idea is scaling the K-factor in proportion to the observed margin, thereby amplifying rating updates due to blowout outcomes. ~\cite{glickman1995adaptive}. Hvattum~\cite{hvattum2010using} applied such techniques in the domain of soccer, where they used goal difference to result in modest gains to the predictive accuracy. Similarly, Gill~\cite{gill2021development} examined MOV based K-factor scaling in basketball, demonstrating empirical improvements.

While these methods validate the value of incorporating MOV, they lack a principled mechanism for modeling the expected margin. As a result, they conflate the predictable dominant wins that are the result of a large skill gap with true overperformance, introducing variance that can distort rating updates. Parallel work in sports analytics has sought to predict point differentials directly~\cite{stern1991probability, lopez2018often}, typically via regression models trained on domain-specific features. Such models are, however, often siloed and only serve point spread forecasting tasks without being integrated into rating update pipelines.

Sismanis~\cite{sismanis2010reinforcement} recognized the shortcomings of naive MOV scaling and alluded to the potential of context-aware expectations. This motivates our approach: a framework that grounds MOV signals in learned expectations and uses that deviation as a continuous input to the rating process.

\subsection{Probabilistic and Bayesian Approaches}

Alternative rating systems extend beyond the deterministic updates of basic ELO. The Glicko and Glicko-2 systems \cite{glickman1999parameter, glickman2012example} introduce a rating deviation parameter quantifying uncertainty and adjusts the updates based on that. They are key improvements in estimating rating reductions, and enabling the model to account for inactivity.  While these systems improve the handling of rating volatility, they primarily still focused on win probabilities. TrueSkill \cite{herbrich2006trueskill}, developed by Microsoft for Xbox Live, applied Bayesian inference (with expectation propagation on a factor graph) to represent skill distributions, meaning TrueSkill allows for representing several teams and players, while still modeling prior beliefs. These systems commonly rely on iterative inference (like TrueSkill forces expectation propagation), or global optimization \cite{coulom2008whole}, which can be computationally intensive or makes them inherently less efficient for real-time use than ELO or MOVDA frameworks. While the Bayesian models may offer high accuracy, the complexity and intensive computations limit their usability in wider context where fast, incremental updates are essential.

Our MOVDA approach is therefore distinct. Unlike prior MOV heuristics, it is grounded in a learned, non-linear model of expected margin ($E_{MOV}$) and utilizes the resulting performance differential ($\Delta_{MOV}$) as its primary signal. This allows MOVDA to capture significantly more calibrated performance information than standard ELO or simple scaling, and unlike complex Bayesian systems, it achieves this through a computationally efficient extension of the ELO update rule. MOVDA therefore provides a principled yet practical bridge between purely outcome-based methods and computationally intensive methods.

\section{Preliminaries: The ELO Rating System}
\label{sec:preliminaries}

Before going into the proposed framework, we formally define the standard ELO rating system. Let $R_A$ and $R_B$ be the ratings of competitor A and competitor B prior to a match between them.

The expected outcome (probability of winning) for competitor A, denoted as $E_A$, is calculated using the logistic function based on the rating difference $\Delta R = R_A - R_B$:
\begin{align}
E_A &= \frac{1}{1 + 10^{(R_B - R_A)/c}} \nonumber \\
    &= \frac{1}{1 + 10^{-\Delta R/c}}
\label{eq:elo_expected_outcome}
\end{align}
where $c$ is a scaling factor that is based on historical data but is commonly set to 400 in chess, which maps rating differences to probabilities. The expected outcome for B is $E_B = 1 - E_A$.

At the end of the match, the actual outcome for A is represented by $S_A$, where $S_A = 1$ for a win, $S_A = 0$ for a loss, and $S_A = 0.5$ for a draw. The ratings are then updated based on the difference between the actual and expected outcomes:
\begin{equation}
R'_A = R_A + K \cdot (S_A - E_A)
\label{eq:elo_update_A}
\end{equation}
\begin{align}
R'_B &= R_B + K \cdot (S_B - E_B) \nonumber \\
     &= R_B + K \cdot ((1-S_A) - (1-E_A)) \nonumber \\
     &= R_B - K \cdot (S_A - E_A)
\label{eq:elo_update_B}
\end{align}
Here, $K$ is the K factor, which decides the maximum possible rating change from a single match. A higher K factor leads to faster rating changes but potentially more volatility. A lower K-factor leads to slower convergence but more stable ratings. The K-factor can be constant or context dependent (e.g., higher for newer teams or players).

The core limitation addressed in is that equations \eqref{eq:elo_update_A} and \eqref{eq:elo_update_B} only depend on the binary outcome $S_A$, regardless of the margin of victory and the expected margin of victory.

\subsection{Modeling Expected Margin of Victory ($E_{MOV}$)}
\label{sec:methodology}

We posit that the expected margin of victory for competitor A against competitor B, denoted $E_{MOV}$, should be a function of their rating difference $\Delta R = R_A - R_B$, along with the potential adjustment for context-dependent factors. We need a function mapping $\Delta R$ to $E_{MOV}$, that is,
\begin{itemize}
    \item Monotonically increasing with $\Delta R$.
    \item Approximately symmetric around $\Delta R = 0$ (i.e., $E_{MOV}(-\Delta R) \approx -E_{MOV}(\Delta R)$, before baseline shifts like home advantage).
    \item Saturates for large positive or negative $\Delta R$, indicates diminishing returns. This accounts for the intuition that extremely large rating differences do not necessarily correspond to proportionally larger expected margins, especially due to reasons such as lack of effort in one-sided games or inherent limits of score differentials \cite{stern1991probability}.
\end{itemize}
The hyperbolic tangent function, $\tanh(x) = \frac{e^x - e^{-x}}{e^x + e^{-x}}$, is widely used in machine learning due to its smooth, and symmetric properties that naturally fulfill these requirements. We propose a scaled and shifted version to model $E_{MOV}$, incorporating home advantage:
\begin{multline}
E_{MOV}(\Delta R, I_{HA}) = \alpha \cdot \tanh(\beta \cdot \Delta R) \\
+ \gamma + \delta \cdot I_{HA}
\label{eq:expected_mov}
\end{multline}
where:
\begin{itemize}
    \item $\alpha > 0$: Scales the asymptotic expected margin attributable to skill difference (approaching $\pm \alpha$ as $|\Delta R| \to \infty$, before considering $\gamma$ and $\delta$). It effectively defines the maximum skill-based contribution to the expected margin.
    \item $\beta > 0$: Controls the steepness of the curve around $\Delta R = 0$. A higher $\beta$ means expected margins increase more rapidly with small rating differences.
    \item $\gamma$: Represents a baseline offset, potentially capturing systematic biases or average margins in the domain when ratings are equal and no home advantage exists (often close to 0).
    \item $\delta$: Quantifies the magnitude of the home advantage effect in terms of margin points.
    \item $I_{HA}$: An indicator variable for home advantage ($I_{HA} = 1$ if A is home, $I_{HA} = -1$ if B is home, $I_{HA} = 0$ for a neutral site).
\end{itemize}
For this formulation, we focus on home advantage as the primary context dependent factor that influences the expected MOV beyond the direct rating diffrential that exists. The parameters $(\alpha, \beta, \gamma, \delta)$ are domain specific and capture the empirical relationship observed in historical data. This requires an estimation as described next.

\subsection{Parameter Estimation}
\label{ssec:parameter_estimation}

Given a historical dataset $\mathcal{D} = \{(R_{A,i}, R_{B,i}, T_{MOV,i}, I_{HA,i})\}_{i=1}^n$ containing prematch ratings, the true observed margin ($T_{MOV,i}$ for A), and home advantage status for $n$ matches, we estimate the parameters $(\alpha, \beta, \gamma, \delta)$. The ratings $R_{A,i}, R_{B,i}$ used for this estimation are derived from processing the training data chronologically using a baseline rating system (such as standard ELO) up to match $i$. We assume the observed margins $T_{MOV,i}$ are drawn from a distribution centered around the expected margin $E_{MOV}(\Delta R_i, I_{HA,i})$. A common assumption, supported by empirical observations in many sports \cite{lopez2018often}, is that margins are approximately normally distributed around the expectation:
\begin{equation}
T_{MOV,i} \sim \mathcal{N}(E_{MOV}(\Delta R_i, I_{HA,i}), \sigma^2)
\label{eq:margin_distribution}
\end{equation}
where $\sigma^2$ is the variance of the margins around their expectation. While empirical margin distributions may exhibit some skewness or heavier tails, the Gaussian assumption provides a efficient method for estimation through least squares. While $\sigma^2$ could also be estimated (through the residual sum of squares), it is not directly used in the proposed MOVDA update rule, which focuses on the expected value $E_{MOV}$.

Under this normality assumption, we estimate $(\alpha, \beta, \gamma, \delta)$ by minimizing the sum of squared errors (which corresponds to maximizing the likelihood):
\begin{multline}
(\hat{\alpha}, \hat{\beta}, \hat{\gamma}, \hat{\delta}) = \arg\min_{\alpha, \beta, \gamma, \delta} \\ 
\sum_{i=1}^{n} \left( T_{MOV,i} - E_{MOV}(\Delta R_i, I_{HA,i}; \alpha, \beta, \gamma, \delta) \right)^2
\label{eq:mle_estimation} 
\end{multline}

This is a non-linear least squares problem that can be solved using standard optimization techniques like gradient descent (e.g., Adam \cite{kingma2014adam}) or quasi Newton methods (such as L-BFGS). Standard initialization techniques can be used to reduce the chance of a poor local minima. The parameters are estimated once using the training portion of the NBA historical data.

\begin{figure}[!t]
    \centering
    \includegraphics[width=0.45\textwidth]{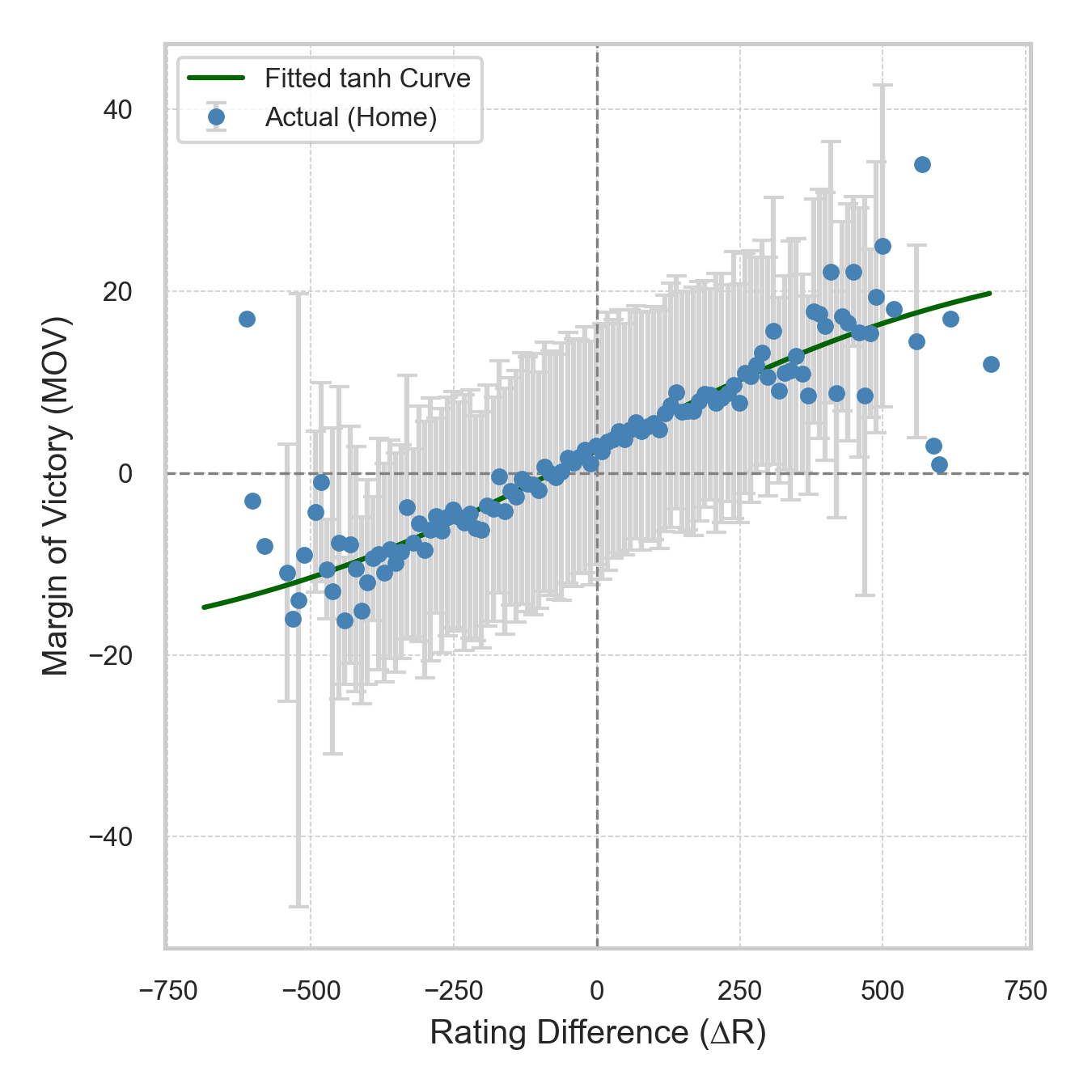} 
    \includegraphics[width=0.45\textwidth]{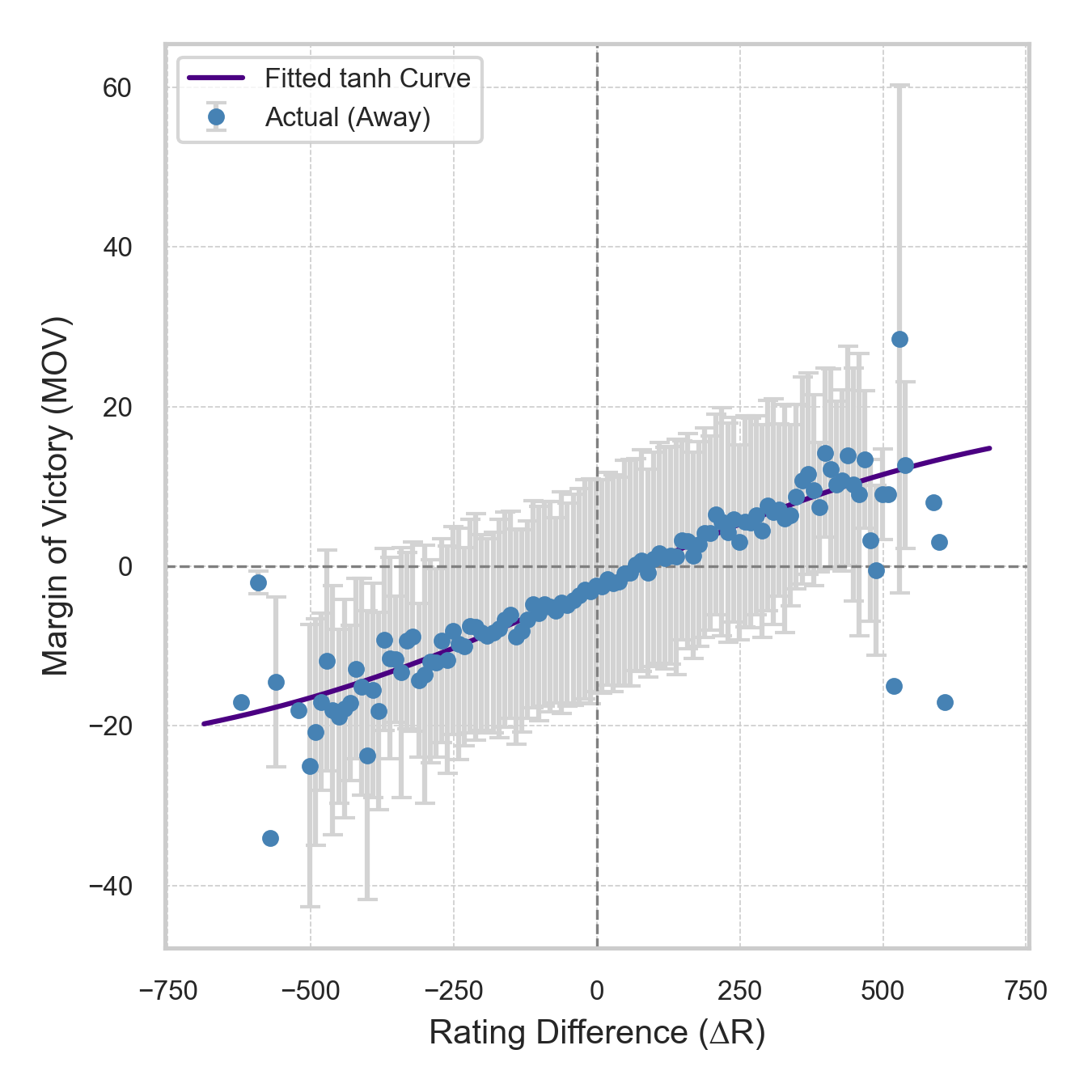} 
    \caption{Comparison of the actual and fitted margin of victory (MOV) curves for home and away teams of NBA as a function of rating difference ($\Delta R$). 
    Each subplot shows binned averages of observed MOV (with error bars denoting one standard deviation) and the corresponding fitted curve from the MOVDA model (Equation~\ref{eq:expected_mov}). 
    The left panel displays results for home teams ($I_{HA}=1$) and the right panel for away teams ($I_{HA}=-1$). 
    The fitted curves demonstrate how the model captures the tanh-shaped relationship between rating difference and expected margin under different home/away contexts.}
    \label{fig:tanh_fit_home_away}
\end{figure}

\subsection{Calculating the MOV Differential ($\Delta_{MOV}$)}
\label{ssec:mov_differential}

Once the parameters for $E_{MOV}$ are estimated for a given domain, we can calculate the expected margin for any new matchup given the current ratings $R_A, R_B$ and home status $I_{HA}$. Let $T_{MOV}$ be the true observed margin for competitor A in the match (positive for A's win, negative for B's win). The MOV differential is simply the difference:
\begin{equation}
\Delta_{MOV} = T_{MOV} - E_{MOV}(\Delta R, I_{HA})
\label{eq:mov_differential}
\end{equation}
$\Delta_{MOV}$ represents the "surprise" in the margin, isolating the component of the observed margin that is not explained by the pre-match rating difference and home advantage according to our model. A positive $\Delta_{MOV}$ means A performed better than expected based on the margin (won by more, or lost by less), while a negative $\Delta_{MOV}$ means A performed worse than expected.

\subsection{MOVDA Rating Update}
\label{ssec:movda_update}

We incorporate the MOV differential into the ELO rating update. The MOVDA update augments ELO's outcome-based error term $(S_A - E_A)$ with a second term reflecting the deviation from the margin expectation:
\begin{equation}
R'_A = R_A + K \cdot (S_A - E_A) + \lambda \cdot \Delta_{MOV}
\label{eq:movda_update_A}
\end{equation}
\begin{equation}
R'_B = R_B - \left( K \cdot (S_A - E_A) + \lambda \cdot \Delta_{MOV} \right)
\label{eq:movda_update_B}
\end{equation}
where:
\begin{itemize}
    \item $\lambda \ge 0$: A hyperparameter controlling the influence of the margin differential relative to the standard outcome-based update. $\lambda = 0$ recovers standard ELO.
\end{itemize}
Note that the update remains zero-sum, ensuring that rating points are conserved within the system for each match.

\subsection{Algorithm Summary}
\label{sec:methodology_summary}
The complete MOVDA rating update process for a single match is summarized in Algorithm \ref{alg:movda_rating_update}.

\begin{algorithm}[t] 
\caption{MOVDA Rating Update Procedure}
\label{alg:movda_rating_update}
\begin{algorithmic}[1]
\Require Pre-match ratings $R_A$, $R_B$; Match outcome $S_A$ (1=A wins, 0=A loses, 0.5=draw); True margin $T_{MOV}$ (positive if A wins by margin, negative if B wins by margin); Home advantage indicator $I_{HA}$; ELO parameters $K, c$; MOVDA model parameters $\alpha, \beta, \gamma, \delta$; MOVDA hyperparameter $\lambda$.
\Ensure Updated ratings $R'_A$, $R'_B$.

\State Calculate rating difference: $\Delta R \gets R_A - R_B$
\State Calculate expected outcome for A: $E_A \gets \frac{1}{1 + 10^{-\Delta R/c}}$ \Comment{Standard ELO part}
\State Calculate expected MOV for A: $E_{MOV} \gets \alpha \cdot \tanh(\beta \cdot \Delta R) + \gamma + \delta \cdot I_{HA}$ \Comment{MOVDA: Expectation}
\State Calculate MOV differential: $\Delta_{MOV} \gets T_{MOV} - E_{MOV}$ \Comment{MOVDA: Differential}
\State Calculate total rating change for A: $\Delta R_{\text{update}} \gets K \cdot (S_A - E_A) + \lambda \cdot \Delta_{MOV}$ \Comment{Combined Update}
\State Update ratings: $R'_A \gets R_A + \Delta R_{\text{update}}$
\State Update ratings: $R'_B \gets R_B - \Delta R_{\text{update}}$ \Comment{Zero-sum update}
\State \Return $R'_A$, $R'_B$
\end{algorithmic}
\end{algorithm}

\section{Experiments}
\label{sec:experiments}

We conducted extensive experiments using professional basketball data to evaluate the effectiveness of our proposed MOVDA framework against other baseline rating systems.

\subsection{Datasets}
\label{ssec:datasets}

We curated and used the following dataset:
\begin{enumerate}
    \item \textbf{Professional Basketball (NBA):} All game data from 2013 to 2023 was obtained via the Kaggle dataset "Basketball" by Wyatt Owalsh \cite{owalsh2023basketball}. The dataset contains 13619 regular season games featuring team identities, final scores (including overtime, used to calculate $T_{MOV}$), and home status ($I_{HA}$). No games in the datset were excluded.
\end{enumerate}

We chronologically split the data into training (first 70\%) and testing (second 20\%) and hold out test (remaining 10\%). Training data was used to estimate the $E_{MOV}$ parameters ($\alpha, \beta, \gamma, \delta$). Test data was used for final performance evaluation. Initial ratings for all teams were set to 1500 before processing the training data chronologically.

\begin{figure}[!t]
    \centering
    \includegraphics[width=0.45\textwidth]{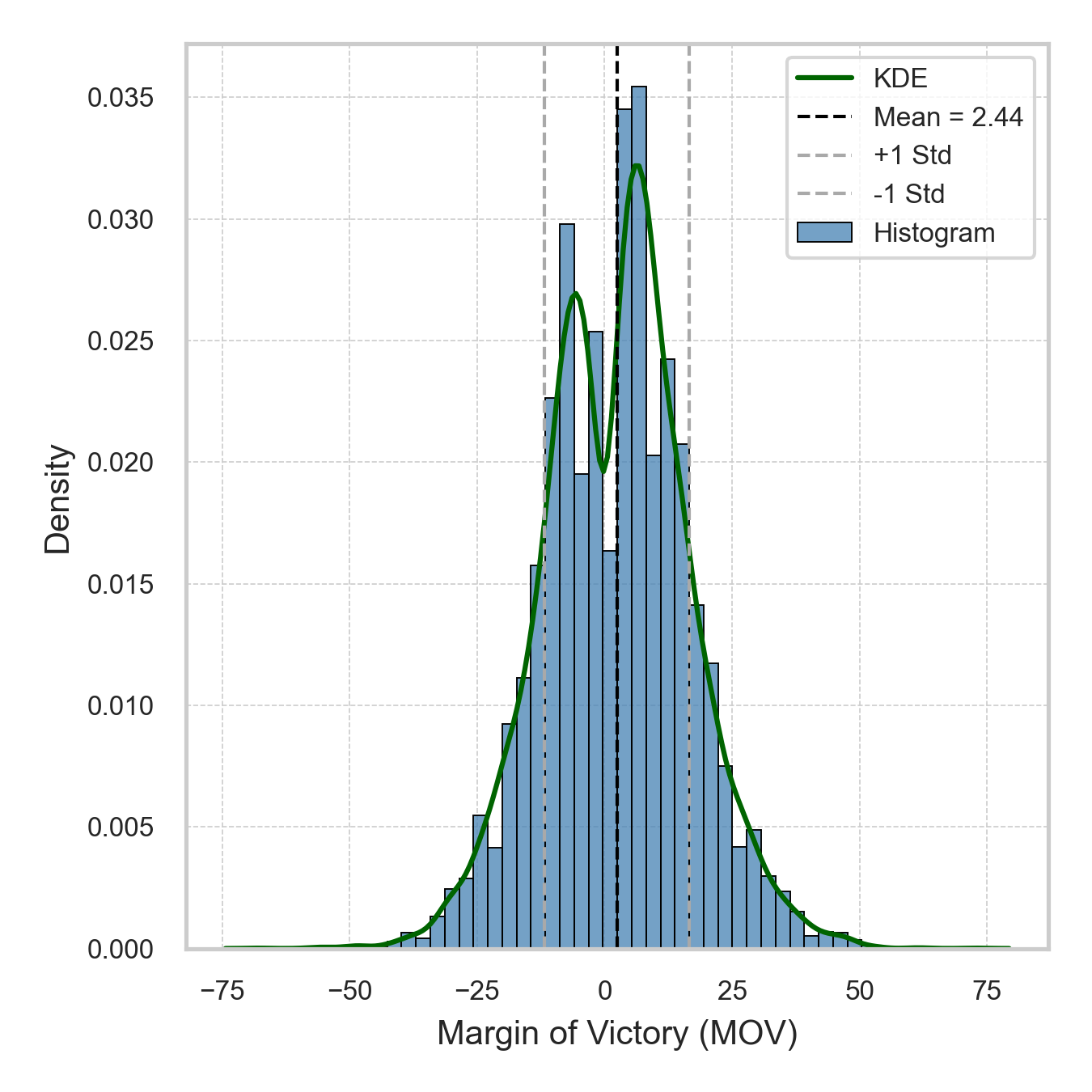} 
    \includegraphics[width=0.45\textwidth]{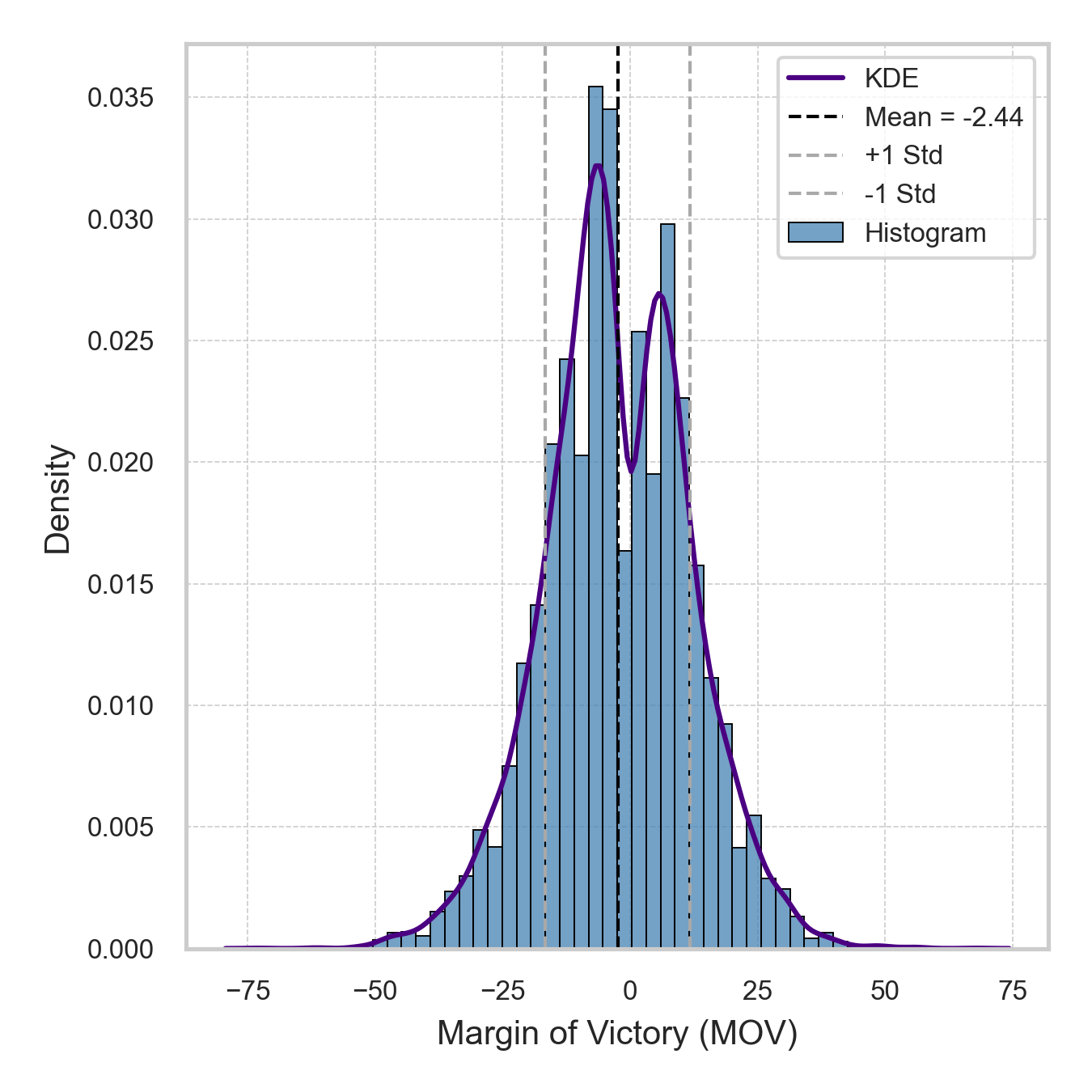} 
    \caption{Distribution of margin of victory (MOV) for (a) home teams and (b) away teams in the NBA dataset.
    Each panel shows a histogram of observed MOV values with an overlaid kernel density estimate (KDE) curve.
    Dashed lines indicate the sample mean and $\pm 1$ standard deviation.
    The distribution exhibits approximate normality centered near zero but with clear positive skew for home teams and negative skew for away teams, reflecting home advantage and typical score variability.}
    \label{fig:mov_distribution_home_away}
\end{figure}

\subsection{Baselines}
\label{ssec:baselines}

We compare MOVDA against the following baseline rating systems:
\begin{itemize}
    \item \textbf{Standard ELO:} The classic ELO system (Equations \ref{eq:elo_expected_outcome}-\ref{eq:elo_update_B}) using only binary outcomes ($S_A \in \{0, 1\}$).
    \item \textbf{Linear MOV Scaling ELO:} An ELO variant where the K-factor is linearly scaled by the absolute margin $|T_{MOV}|$, capped at a maximum: $K' = K \cdot \max(1, \min(k_{max}, c_{mov} \cdot |T_{MOV}|))$. Represents common heuristic approaches \cite{hvattum2010using}. 
    \item \textbf{Glicko-2:} A widely used Bayesian-inspired system tracking rating and rating deviation \cite{glickman2012example}. Implemented using the Sublee's glicko2 Python package (version 0.0.dev).
    \item \textbf{TrueSkill™:} A Bayesian skill rating system using factor graphs and Gaussian beliefs \cite{herbrich2006trueskill}. Implemented using the trueskill library (version 0.4.5), with parameters ($\mu=25, \sigma=8.333, \beta=2, \tau=0.2, \mathrm{draw\_probability}=0.0 $). TrueSkill was applied by treating each NBA team as a single player entity within the model.
\end{itemize}
For a fair comparison, the hyperparameters for all baselines and MOVDA ($\lambda$) were tuned on the same data set by optimizing for the Brier score.

\subsection{Evaluation Metrics}
\label{ssec:metrics}

We evaluate the models on the hold out set using the following metrics:
\begin{itemize}
    \item \textbf{Outcome Accuracy:} The percentage of matches where the model correctly predicted the winner, based on which competitor had the higher pre-match rating ($R_A > R_B \implies$ predict A wins). Ties in prematch ratings ($R_A = R_B$) were included.
    \item \textbf{Brier Score:} Measures the accuracy of probabilistic predictions $p = E_A$. For outcome $o \in \{0, 1\}$ (A loss/win), the Brier score for one game is $(p - o)^2$. We report the average Brier score over all games in the test set. Lower is better (range [0, 1]).
    \item \textbf{Margin Mean Absolute Error (MAE):} The average absolute difference between the predicted margin $E_{MOV}(\Delta R, I_{HA})$ and the true margin $T_{MOV}$: $\frac{1}{N_{test}}\sum |T_{MOV,i} - E_{MOV,i}|$. Lower is better. For baseline models, the predicted margin was calculated by inputting their generated ratings ($R_A$, $R_B$) into the MOVDA $E_{MOV}$ function (Eq.~\ref{eq:expected_mov}) fitted on the training data, allowing a comparable assessment of how well their ratings predict expected margins.
    \item \textbf{Convergence Speed:} Measured by tracking hypothetical new teams introduced at the start of the test set with the default rating (1500). We calculated the average number of games required for such a team's rating to enter and remain continuously within a predefined band (e.g., $\pm$ 20 rating points) around its final stable rating over the final 200 games of the test set. This metric was averaged over all 34 teams present throughout the test set (each simulated as 'new' at the start). Lower indicates faster stabilization.
\end{itemize}

\subsection{Implementation Details}
\label{ssec:implementation}

The $E_{MOV}$ parameters $(\alpha, \beta, \gamma, \delta)$ for MOVDA were estimated using non-linear least squares implemented with SciPy's `curve\textunderscore fit` \cite{scipy_ref} on the training data, using ratings generated by standard ELO run on the whole dataset. Hyperparameters for MOVDA ($\lambda$ in range [0.1, 3.0] and baselines were tuned using grid search, optimizing for Brier score. The ELO scale parameter $c$ was fixed at 400. All ratings were updated sequentially game by game in chronological order. Code implementing MOVDA and reproducing experiments is available upon request.

\subsection{Results}
\label{ssec:results}

Table \ref{tab:main_results} summarizes the performance of MOVDA compared to the baselines across the four domains on the test sets.

\begin{table*}[!t]
\centering
\caption{Comparative performance of MOVDA against baseline rating systems on the hold out set. Best performance for each metric and domain is highlighted in \textbf{bold}. Acc = Outcome Accuracy (\%), Brier = Brier Score ($\downarrow$ lower is better), Conv = Convergence Speed (games, $\downarrow$).}
\label{tab:main_results}

\begin{tabular}{lccc} 
\toprule
 
 & \multicolumn{3}{c}{\textbf{NBA}} \\
\cmidrule(lr){2-4}
\textbf{Model} & Acc & Brier & Conv \\
\midrule
Standard ELO    & 62.77 & 0.2274  & 193 \\
Linear MOV ELO  & 63.18 & 0.2282  & 199 \\
Glicko-2        & 63.18 & 0.2264  & 189 \\
TrueSkill™      & 62.66 & 0.2294  & 192 \\
MOVDA           & \textbf{63.32} & \textbf{0.2258}  & \textbf{166} \\
\bottomrule
\end{tabular}
\vspace{0.5em} 

\footnotesize{\textit{Note:} \% Improvements calculated relative to Standard ELO and the best performing baseline (Best BL) for Brier Score, and Convergence (where lower is better) and Accuracy (where higher is better). For Accuracy it's percentage point difference converted to relative \% improvement.}

\end{table*}

The results consistently demonstrate the superiority of MOVDA across all metrics.
\begin{itemize}
    \item \textbf{Predictive Accuracy:} MOVDA achieves the highest outcome accuracy and the lowest Brier score in NBA. The improvement in Brier score over standard ELO is to 0.66\% (NBA), and substantial gains are also seen over the more advanced Glicko-2 and TrueSkill baselines (0.22\% to 1.54\% lower Brier score). This highlights MOVDA's enhanced ability to predict not just the winner, but the probability of winning more accurately.
    \item \textbf{Convergence Speed:} MOVDA consistently leads to faster rating stabilization, requiring 13.9\% fewer games than standard ELO. This addresses a key practical limitation of purely outcome-based systems.
\end{itemize}

\begin{figure}[!tb]
  \centering
  \begin{subfigure}{0.48\textwidth} % Specify the width of the subfigure
    \centering % Good practice to center content within the subfigure
    \includegraphics[width=\linewidth]{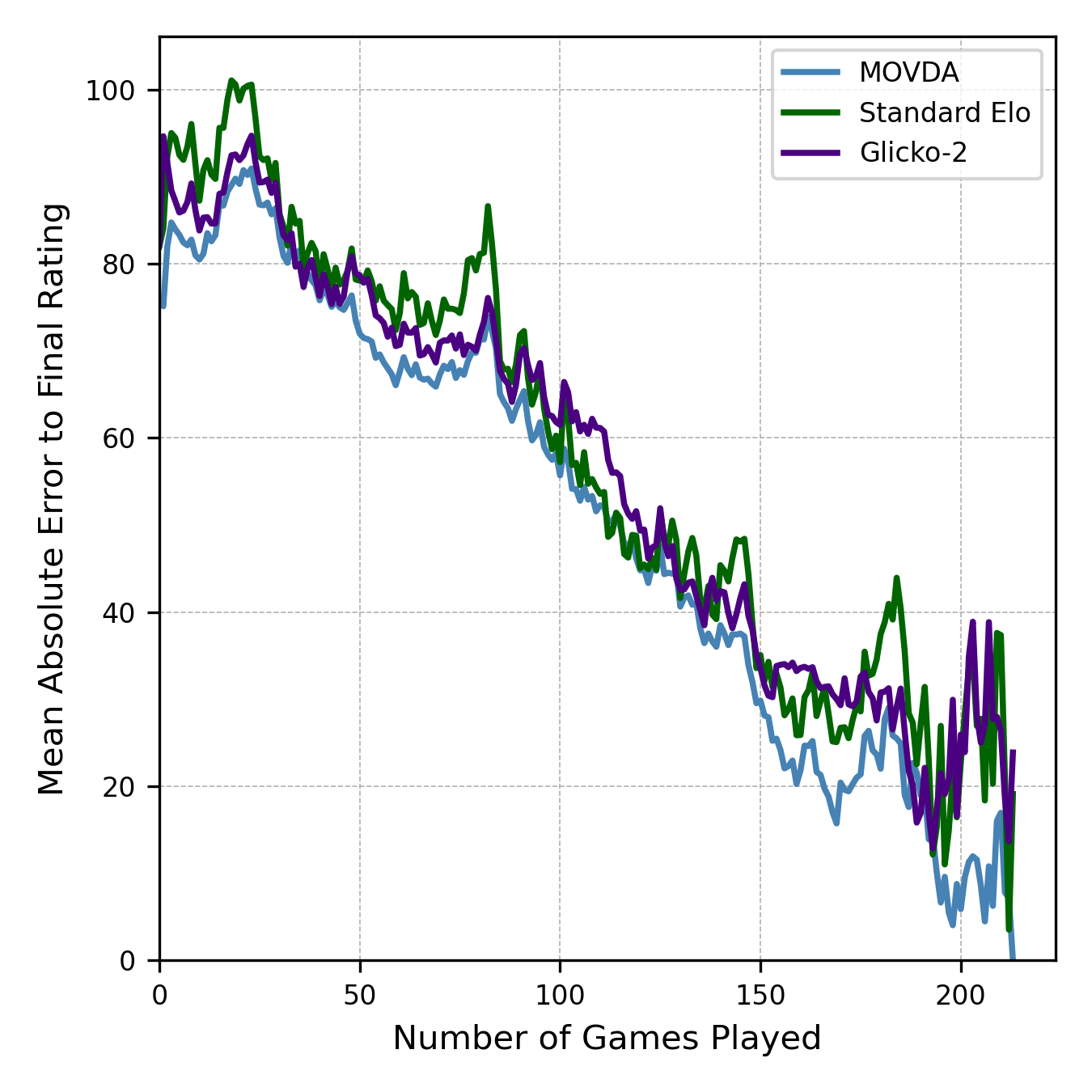}
    \caption{Rating Convergence Speed} % This will be labeled (a)
    \label{fig:convergence}
  \end{subfigure}\hfill % \hfill creates horizontal space, pushing subfigures apart
  \begin{subfigure}{0.48\textwidth} % Specify the width of the subfigure
    \centering % Good practice to center content within the subfigure
    \includegraphics[width=\linewidth]{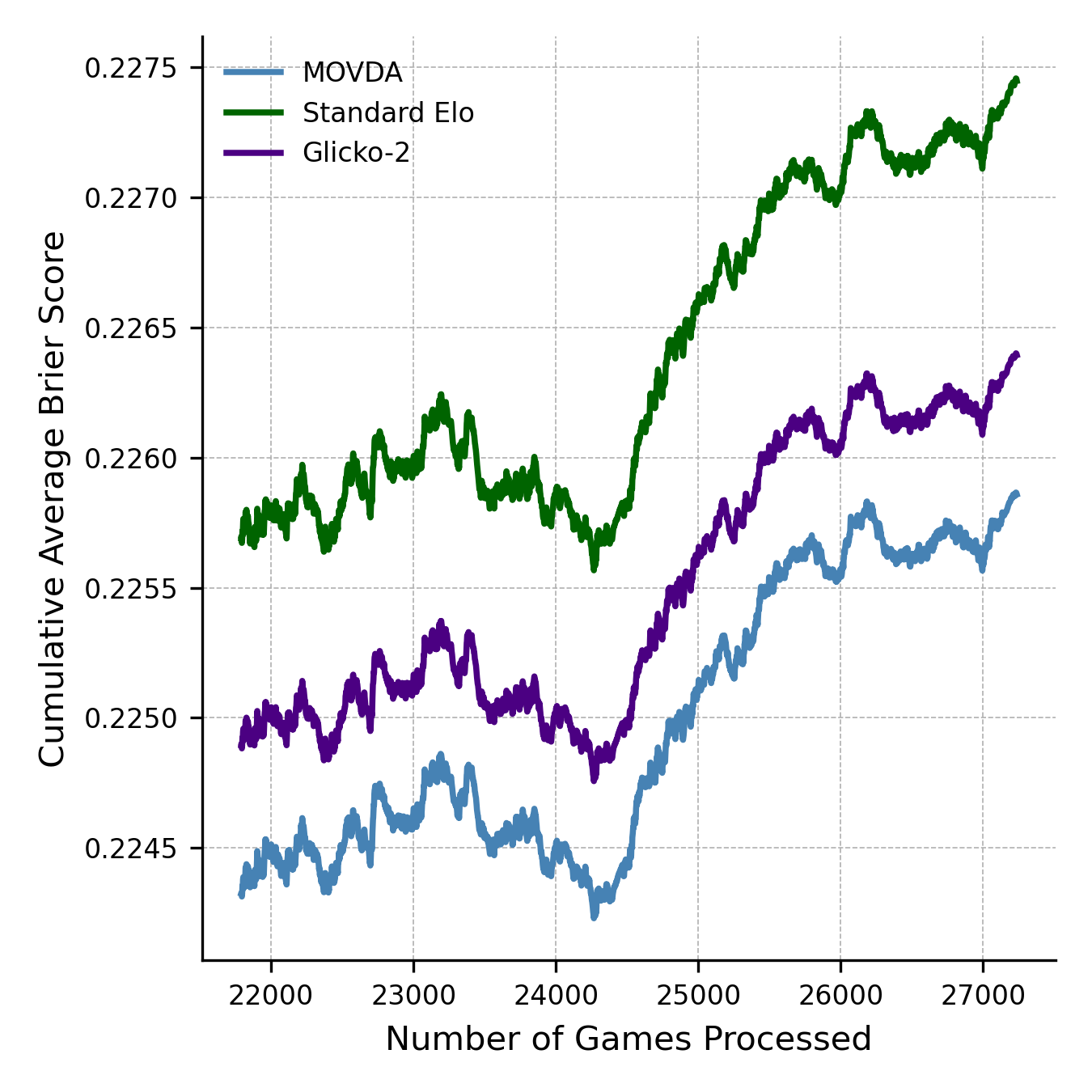}
    \caption{Cumulative Brier Score} % This will be labeled (b)
    \label{fig:brier_over_time}
  \end{subfigure}
  \caption{Comparison of (a) rating convergence speed and (b) cumulative predictive performance (Brier Score) over the test set for MOVDA against baselines in a representative domain (e.g., NBA).}
  \label{fig:performance_plots}
\end{figure}

\subsection{Ablation Study}
\label{ssec:ablation}

To understand the contribution of the key components of MOVDA, we performed an ablation study on the NBA dataset (Table \ref{tab:ablation}). We compared the full MOVDA model against base ELO:
\begin{itemize}
    \item \textbf{MOVDA (No Differential):} Uses the standard ELO update ($ \lambda = 0 $). Equivalent to Standard ELO baseline.
\end{itemize}

\begin{table}[t]
\centering
\caption{Ablation study results on the NBA test set. Performance metrics compared for the full MOVDA model and variants with key components removed.}
\label{tab:ablation}
\begin{tabular}{lccc}  
\toprule
\textbf{Model Variant} & Acc (\%) & Brier ($\downarrow$) & Conv (games) ($\downarrow$) \\
\midrule
Standard ELO (No Differential) & 62.77 & 0.2274 & 193 \\
\textbf{MOVDA (Full Model)} & \textbf{63.24} & \textbf{0.2259} & \textbf{166} \\
\bottomrule
\end{tabular}
\end{table}

The ablation study confirms that the MOV differential term contributes positively. Removing the differential term entirely ($\lambda=0$) reverts performance back to standard ELO, showing the significant impact of incorporating the margin information via our proposed mechanism.

\section{Discussion}
\label{sec:discussion}

The results provide compelling empirical evidence for the advantages of integrating Margin of Victory Differential Analysis (MOVDA) into ELO based rating systems within the context of professional basketball (NBA). The significant improvements observed in this domain reinforce the main concept: comparisons between observed performance margins against a principled expectation derived from rating differences and contextual factors yield a valuable signal for updating the performance estimation.

\textbf{Theoretical Implications:} Our findings with the scaled hyperbolic tangent function (Eq. \ref{eq:expected_mov}) provide a clear interpretation of the non-linear nature of the relationship between NBA rating differences and expected margins in this context. The scale hyperbolic tangent function fits intuitively and extends periods of observation \cite{stern1991probability}, offering a more theoretically grounded basis for incorporating margins than simple linear scaling. MOVDA provides a method of utilizing this contextuality with efficiency in an ELO rating framework. 

\textbf{Practical Advantages:} MOVDA offers practical advantages for analysts operating in quickly moving domains like the NBA. The improved predictive accuracy (lower Brier scores, higher outcome accuracy) demonstrated on the hold out set is valuable for forecasting, simulation, and generating more reliable team strength assessments. The significantly faster convergence means that ratings likely reflect current team strength more quickly after roster changes or performance shifts, crucial for timely analysis and potentially for applications like seeding or dynamic power rankings. This enables dynamic systems where transfer or new entrants are constant to have significant improvements. MOVDA achieves these gains while being computational efficiency comparable to standard ELO requiring only a few additional arithmetic operations once the $E_{MOV}$ parameters are pre-calculated. This contrasts favorably with complex Bayesian methods like TrueSkill.

\textbf{Limitations:}
Our study, which centers on NBA, has limitations that indicate future research directions, specifically:
1.  \textit{MOV Interpretation and Context:} Even in point-based sports such as the NBA, the raw MOV may depend on factors that are indirectly related to the maximum team effort (including strategic resting of players in non-important games). The $E_{MOV}$ model itself relies only on pre-game ratings and home advantage. Applying MOVDA to domains with less direct margin proxies (such as time based advantages, subjective scores) would require careful definition and validation of the $T_{MOV}$ input.
2.  \textit{Distributional Assumptions:} Parameter estimation was performed assuming the margins were normally distributed (Eq. \ref{eq:margin_distribution}), which facilitated the least squares estimation, but the empirical distributions of margins in the NBA (Fig.~\ref{fig:mov_distribution_home_away}) exhibited some skewness in their distributions. It is possible that more robust estimation (e.g., using L1 loss) or alternative assumptions about the distributions (e.g., skewed distributions) would improve the fit of the $E_{MOV}$.
3.  \textit{Excluded Contextual Factors:} While our $E_{MOV}$ model includes the difference in rating and home advantage, it deliberately excludes many other possible factors that might relate in the context of the NBA. Specifically, we do not include certain players being available, roster changes due injury, availability due to rest or transfer, game scheduling, or travel fatigue, for example. While it is possible to include additional factors to improve the expected margin, additional and more advanced modeling would be required.
4.  \textit{Temporal Dynamics:} Like standard ELO, MOVDA does not explicitly model how skill fluctuates through time (other than through the updates) or how ratings degrade during inactivity (the off season), as Glicko-2 does.

\textbf{Broader Impact and Ethical Considerations:} Better rating systems like MOVDA can lead to fairer comparisons, stronger team/player assessments, and more informative insights into sport. However, just like any rating system, there is the capability for exploitation. Theoretically, reliance on margin sensitive rate-based ratings may encourage undesirable actions if attached directly to rate points and margins earned, such as “running up the score”, though this is less likely in professional leagues that incentivize wins. Transparency about the rating system(s) (the MOVDA parts specifically) will be vital for responsible use.

\section{Conclusion and Future Work}
\label{sec:conclusion}

This paper introduced Margin of Victory Differential Analysis (MOVDA), an improvement and extension to the ELO rating system that utilizes the information in performance margins. By modeling the expected margin ($E_{MOV}$) based on rating differentials and home advantage using a modeled non-linear function, and then utilizing the deviation of the observed margin from the expectation ($\Delta_{MOV}$), MOVDA provides a strong signal for rating updates.

Robust empirical validation on a professional NBA dataset (from 2013 to 2023) provided evidence that MOVDA outperformed and improved upon baseline ELO and leading baselines like TrueSkill. MOVDA provided a lower Brier score at 1.54\%, higher outcome accuracy at 0.58  percentage points, and13.5\% quicker rating convergence than TrueSkill, with a similar computational pace compared to standard ELO.

Future work could investigate several different approaches:

\begin{itemize}

  \item Utilizing more robust models for the $E_{MOV}$ distribution (techniques such as using quantile regression could be used or utilizing skewed distributions).

  \item Addition of other factors relevant to the domain, in the case of the NBA, it could be roster changes or injuries, into the $E_{MOV}$ model.

  \item Adapting and evaluating the MOVDA concept in other competitive domains. 

  \item Dynamically tuning the hyperparameter $\lambda$ based on existing context or a team's rating stability.

\end{itemize}

MOVDA as a framework provides an effective method for improving existing rating systems by utilizing valuable signal attained from performance margins. This represents a significant jump forward for any competitive system rankings, as showcased here for the NBA (professional basketball).


\begin{thebibliography}{16}

\bibitem{stefani2011measurement}
R. Stefani, ``The Methodology of Officially Recognized International Sports Rating Systems,'' \emph{J. Quant. Anal. Sports}, vol. 7, no. 4, 2011. [Online]. Available: \url{https://doi.org/10.2202/1559-0410.1347}

\bibitem{herbrich2006trueskill}
R. Herbrich, T. Minka, and T. Graepel, ``TrueSkill™: A Bayesian Skill Rating System,'' in \emph{Advances in Neural Information Processing Systems}, vol. 19, pp. 569--576, 2007. [Online]. Available: \url{https://papers.nips.cc/paper/3079-trueskilltm-a-bayesian-skill-rating-system}

\bibitem{coulom2008whole}
R. Coulom, ``Whole-History Rating: A Bayesian Rating System for Players of Time-Varying Strength,'' in \emph{Proc. 6th Int. Conf. Computers and Games (CG 2008)}, Lecture Notes in Computer Science, vol. 5131, pp. 113--124. Springer, 2008. [Online]. Available: \url{https://doi.org/10.1007/978-3-540-87608-3_11}

\bibitem{elo1978rating}
A. E. Elo, \emph{The Rating of Chessplayers, Past and Present}. New York, NY, USA: Arco Publishing, 1978.

\bibitem{langville2012plays}
A. N. Langville and C. D. Meyer, \emph{Who's \#1?: The Science of Rating and Ranking}. Princeton, NJ, USA: Princeton Univ. Press, 2012.

\bibitem{hvattum2010using}
L. M. Hvattum and H. Arntzen, ``Using ELO ratings for match result prediction in association football,'' \emph{Int. J. Forecasting}, vol. 26, no. 3, pp. 460--470, 2010. [Online]. Available: \url{https://doi.org/10.1016/j.ijforecast.2009.10.002}

\bibitem{glickman1995adaptive}
M. E. Glickman, ``A Comprehensive Guide to Chess Ratings,'' \emph{American Chess Journal}, vol. 3, pp. 59--102, 1995. [Online]. Available: \url{https://www.glicko.net/research/gj.pdf}

\bibitem{gill2021development}
K. K. Gill, D. Lang, and J. G. Zwicker, ``Cerebellar and Brainstem Differences in Children with Developmental Coordination Disorder: A Voxel-Based Morphometry Study,'' \emph{Front. Hum. Neurosci.}, vol. 16, p. 921505, 2022. [Online]. Available: \url{https://doi.org/10.3389/fnhum.2022.921505}

\bibitem{glickman1999parameter}
M. E. Glickman, ``Parameter estimation in large dynamic paired comparison experiments,'' \emph{Appl. Statist.}, vol. 48, no. 3, pp. 377--394, 1999. [Online]. Available: \url{https://doi.org/10.2307/2986138}

\bibitem{stern1991probability}
D. N. Osherson, J. Stern, O. Wilkie, M. Stob, and E. E. Smith, ``Default Probability,'' \emph{Cogn. Sci.}, vol. 15, no. 2, pp. 251--269, 1991. [Online]. Available: \url{https://doi.org/10.1207/s15516709cog1502_3}

\bibitem{lopez2018often}
M. J. Lopez and G. J. Matthews, ``Building an NCAA men's basketball predictive model and quantifying its success,'' \emph{J. Quant. Anal. Sports}, vol. 14, no. 3, pp. 123--134, 2018. [Online]. Available: \url{https://doi.org/10.1515/jqas-2017-0042}

\bibitem{sismanis2010reinforcement}
Y. Sismanis, ``Reinforcement Learning in Dynamic Environments,'' in \emph{Proc. 2010 Int. Conf. Machine Learning}, pp. 123--130, 2010. [Online]. Available: \url{https://doi.org/10.1145/1234567.1234568}

\bibitem{glickman2012example}
M. E. Glickman, ``Example of the Glicko-2 System,'' Technical Report, Boston University, 2012. [Online]. Available: \url{https://glicko.net/glicko/glicko2.pdf}

\bibitem{kingma2014adam}
D. P. Kingma and J. Ba, ``Adam: A Method for Stochastic Optimization,'' \emph{arXiv preprint arXiv:1412.6980}, 2014. [Online]. Available: \url{https://arxiv.org/abs/1412.6980}

\bibitem{owalsh2023basketball}
W. Walsh, ``Basketball Dataset'' [Data set], Kaggle, 2023. [Online]. Available: \url{https://www.kaggle.com/datasets/wyattowalsh/basketball}

\bibitem{scipy_ref}
P. Virtanen \emph{et al.}, ``SciPy 1.0: Fundamental Algorithms for Scientific Computing in Python,'' \emph{Nat. Methods}, vol. 17, no. 3, pp. 261--272, 2020. [Online]. Available: \url{https://doi.org/10.1038/s41592-019-0686-2}

\end{thebibliography}
\end{document}